\documentclass[sigconf]{acmart}
\AtBeginDocument{%
  }

\setcopyright{acmlicensed}
\copyrightyear{2026}
\acmYear{2026}
\setcopyright{cc}
\setcctype{by}
\acmConference[MM '26]{Proceedings of the 34th ACM International Conference on Multimedia}{November 10--14, 2026}{Rio de Janeiro, Brazil}
\acmBooktitle{Proceedings of the 34th ACM International Conference on Multimedia (MM '26), November 10--14, 2026, Rio de Janeiro, Brazil}
\acmDOI{10.1145/3767308.3836343}
\acmISBN{979-8-4007-2213-4/2026/11}

\usepackage{multirow}

\begin{document}

\title{US-VLA: An Ultrasound Vision-Language-Action Model for Embodied Abdominal Scanning}

\author{Cheng Zhang}
\authornote{Both authors contributed equally to this research.}
\affiliation{%
  \institution{Faculty of Computer Science and Technology, Ocean University of China}
  \city{Qingdao}
  \country{China}}
\email{zhangcheng@stu.ouc.edu.cn}

\author{Xingzheng Wu}
\authornotemark[1]
\affiliation{%
  \institution{Faculty of Computer Science and Technology, Ocean University of China}
  \city{Qingdao}
  \country{China}}
\email{wuxingzheng@stu.ouc.edu.cn}

\author{Guihao Yan}
\affiliation{%
  \institution{Faculty of Computer Science and Technology, Ocean University of China}
  \city{Qingdao}
  \country{China}}
\email{yanguihao@stu.ouc.edu.cn}

\author{Xifeng Hu}
\affiliation{%
  \institution{School of Information Science and Engineering, Shandong University}
  \city{Qingdao}
  \country{China}}
\email{202220466@mail.sdu.edu.cn}

\author{Zhi Liu}
\affiliation{%
  \institution{School of Information Science and Engineering, Shandong University}
  \city{Qingdao}
  \country{China}}
\email{liuzhi@sdu.edu.cn}

\author{Mei Wu}
\affiliation{%
  \institution{Department of Ultrasound, the Qilu Second Hospital of Shandong University}
  \city{Jinan}
  \country{China}}
\email{wumei0212@sdu.edu.cn}

\author{Qing Cai}
\correspondingauthor
\affiliation{%
  \institution{Innovation School of Artificial Intelligence, Hefei University of Technology}
  \city{Hefei}
  \country{China}}
\email{caiqing@hfut.edu.cn}

\renewcommand{\shortauthors}{Cheng Zhang et al.}

\begin{abstract}
Artificial intelligence–assisted ultrasound scanning enhances diagnostic reliability and efficiency by providing real-time guidance for standardized image acquisition and reducing operator dependence. However, existing reinforcement learning and learning-assisted ultrasound scanning methods typically rely on carefully designed reward functions or extensive interaction data, which limits their generalization ability and stability across different devices, patient populations, and complex clinical scenarios. To address these challenges, we propose an ultrasound vision-language-action model (US-VLA) for automated ultrasound scanning that explicitly encodes clinical semantic goals and generates sequential probe manipulation actions under real-time ultrasound feedback. In particular, we first design an ultrasound-aware expert fusion module to jointly integrate ultrasound observations with auxiliary contextual information, enabling semantic ultrasound feedback to effectively guide the scanning process. Then, we construct US-VLA-Data, a real-world dataset covering liver and kidney examinations, which includes five clinically defined standard planes and comprises 320 expert scanning trajectories with approximately 80,000 synchronized timesteps. Extensive experiments demonstrate that US-VLA achieves competitive performance in ultrasound probe manipulation tasks, indicating its effectiveness and promising generalization within the evaluated abdominal ultrasound setting. The source code is available at https://github.com/VMVLab/US-VLA.
\end{abstract}




\maketitle

\section{Introduction}

Artificial intelligence–assisted ultrasound scanning plays a critical role in improving the reliability, efficiency, and accessibility of ultrasound-based diagnosis \cite{bib1,bib2,bib39}. Conventional ultrasound examinations rely heavily on operator expertise to acquire clinically meaningful standard planes, resulting in substantial inter-operator variability and inconsistent diagnostic quality \cite{bib3}. By providing real-time perception and action guidance during the scanning process, AI-assisted ultrasound systems facilitate standardized image acquisition, reduce operator workload, and enhance robustness across different devices and patient populations \cite{bib4,bib41,bib42}. Moreover, such systems lay the foundation for intelligent and potentially autonomous ultrasound examinations, extending high-quality diagnostic services to resource-limited clinical settings.

\begin{figure}[t]
\centering
\includegraphics[width=0.48\textwidth]{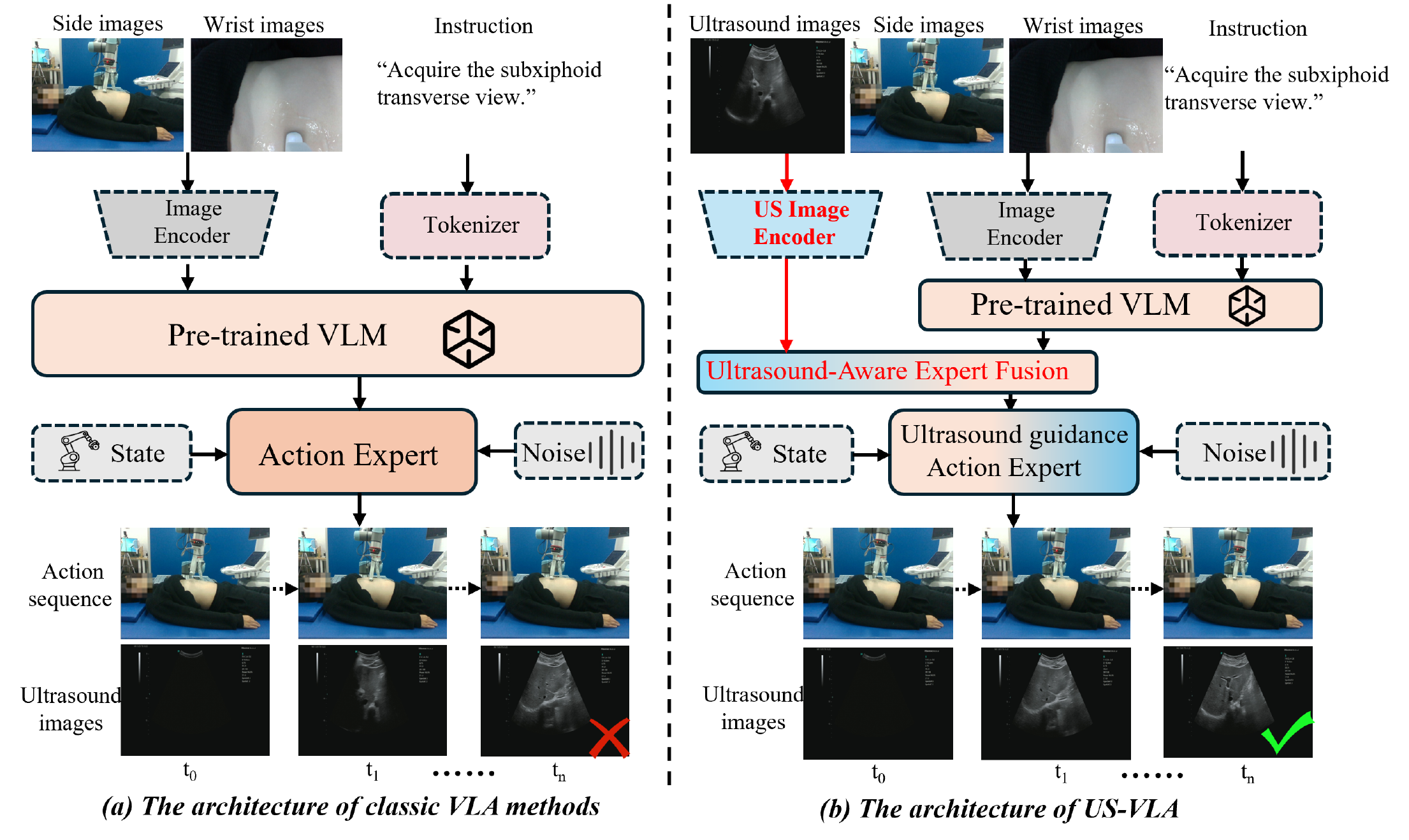}
\caption{Comparison between the architectures of classic VLA methods and the proposed US-VLA. US-VLA encodes ultrasound images and effectively integrates them with vision–language features, facilitating more precise and reliable acquisition of standard ultrasound planes.}
\label{fig1}
\end{figure}

Existing approaches to ultrasound scanning automation mainly fall into reinforcement learning--based \cite{bib5,bib6,bib7,bib40} and learning-based strategies \cite{bib8,bib9,bib10,bib38}. Reinforcement learning methods enable closed-loop probe control through environment interaction, but they heavily rely on carefully designed reward functions, are often trained in simplified settings, and show limited robustness under anatomical variability, patient motion, and real clinical constraints \cite{bib6}. Learning-based methods driven by expert demonstrations can capture clinical scanning priors and achieve stable performance, yet they typically require costly data collection and primarily learn low-level motion patterns without explicitly modeling high-level clinical objectives, which restricts their adaptability to new protocols and scanning targets \cite{bib9}.

Beyond these paradigms, vision--language--action (VLA) models have shown strong performance in robotic environments by aligning language-specified goals with visual perception and sequential action generation \cite{bib11,bib12,bib47,bib48,bib49}. Recent studies have begun to explore VLA paradigms in medical robotics \cite{bib51}, such as RoboNurse-VLA \cite{bib32}, which enables language-conditioned instrument handover via visual perception and large language models. However, existing medical VLA systems mainly focus on object-centric manipulation under external visual observation, where task objectives are defined by tool or object states rather than by imaging-derived semantic criteria. In contrast, ultrasound examination is inherently driven by semantic imaging goals, where success depends on acquiring clinically meaningful standard planes under continuous ultrasound feedback. To date, VLA frameworks have rarely been tailored for imaging-driven diagnostic scanning tasks, motivating the need for ultrasound-aware perception and closed-loop action generation within a unified VLA framework.

Based on the above discussion, we present the VLA-based framework tailored for automated abdominal ultrasound scanning, which explicitly grounds clinical goals into ultrasound-aware perception and closed-loop probe control. As illustrated in Fig.~\ref{fig1}, in contrast to conventional VLA pipelines that primarily rely on external visual observations, US-VLA explicitly incorporates real-time ultrasound feedback and clinical semantic goals to guide probe manipulation. Specifically, US-VLA augments a pre-trained VLM with a dedicated ultrasound image encoder and an ultrasound-aware expert fusion module, which injects task-relevant ultrasound semantics into the action generation pathway and enables an ultrasound-guided action expert to generate sequential probe manipulation actions under closed-loop ultrasound feedback. In addition, we introduce US-VLA-Data, a vision–language–action ultrasound dataset covering liver and kidney examinations, including five clinically defined standard planes, 320 expert scanning trajectories, and approximately 80,000 synchronized timesteps.

The main contributions of this work are summarized as follows:
\begin{itemize}
    \item We present the first vision--language--action (VLA) framework tailored for ultrasound scanning, unifying clinical semantic goals, real-time perception, and probe manipulation within a single model.
    
    \item We construct US-VLA-Data\footnote{\url{https://huggingface.co/datasets/usvla/us_dataset}}, a real-world VLA dataset for ultrasound scanning covering liver and kidney organs and their standard planes, supporting language-conditioned and action-driven learning.
    
    \item Extensive experiments demonstrate that US-VLA consistently outperforms strong baselines and exhibits robust generalization across different organs, scanning targets, and diverse clinical conditions.
\end{itemize}

\section{Related Works}
\subsection{AI-Assisted Ultrasound Scanning}
Ultrasound scanning is highly operator-dependent and requires substantial expertise to acquire clinically meaningful views, motivating extensive research on AI-assisted ultrasound analysis and scanning systems that provide real-time guidance or autonomous probe control \cite{bib13,bib14,bib52}. Early studies mainly explored reinforcement learning and learning-based strategies to guide operators toward target standard planes using ultrasound image feedback, demonstrating feasibility in phantom studies and small-scale clinical trials. These methods typically formulate plane acquisition as a sequential control problem and optimize probe motions through reward-driven exploration or supervised regression of expert trajectories. Subsequent work further extended these paradigms to more realistic and automated settings, including vision-based imitation learning \cite{bib15,bib16}, robot-assisted ultrasound scanning \cite{bib17}, and shared-control systems \cite{bib19,bib20}, where deep models predict probe motion commands from ultrasound observations or fused multimodal inputs. Although these approaches show promising performance, they often rely on task-specific rewards or low-level motion supervision and mainly optimize geometric or appearance cues. Consequently, they lack explicit modeling of clinical semantics and diagnostic intent, limiting robustness, generalization, and adaptability across patients, devices, environments, and examination protocols.


\begin{figure*}[t]
\centering
\includegraphics[width=0.95\textwidth]{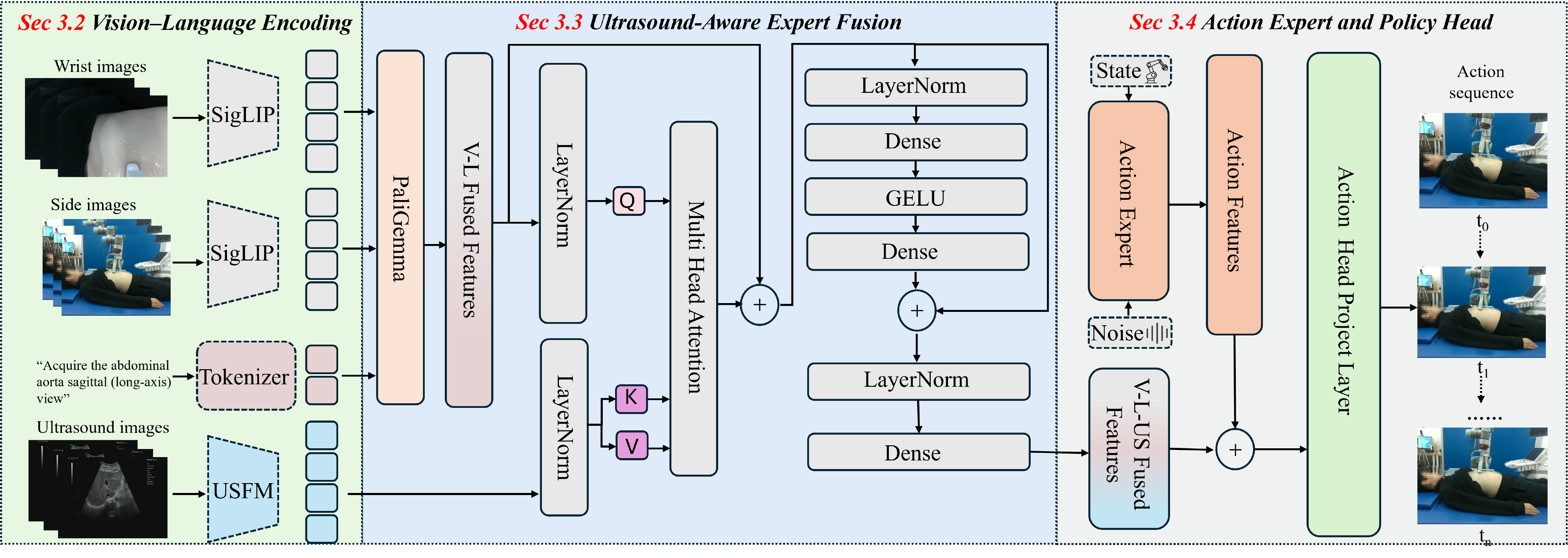}
\caption{The architecture of the US-VLA. The framework consists of three main components: (1) vision–language encoding for aligning clinical semantic goals with external visual observations, (2) ultrasound-aware expert fusion for injecting real-time ultrasound feedback into the decision process, and (3) an action expert and policy head for generating sequential probe manipulation actions under closed-loop ultrasound guidance.}
\label{fig2}
\end{figure*}

\subsection{Vision--Language--Action (VLA) Models}
Vision--language--action (VLA) models map visual observations and language instructions to action sequences through multimodal pretraining, directly grounding semantic goals into embodied control. Existing methods employ end-to-end action decoders \cite{bib21,bib22}, reasoning modules \cite{bib23,bib24,bib36,bib37}, geometric representations \cite{bib25,bib26}, and diffusion policies \cite{bib27,bib28}, achieving strong generalization in navigation and manipulation \cite{bib37}. However, most rely primarily on RGB vision and language, limiting performance in contact-rich and safety-critical scenarios. Recent studies therefore incorporate depth \cite{bib29,bib46}, force \cite{bib30,bib43,bib44,bib45}, or tactile feedback \cite{bib31,bib50} to improve robustness. Extending VLA models to medical environments remains challenging because of perceptual ambiguity, specialized sensing, strict safety requirements, and limited interaction data. Existing medical VLA studies, such as RoboNurse-VLA \cite{bib32}, mainly address visually guided object manipulation in structured surgical settings. In contrast, ultrasound scanning requires continuous imaging feedback and semantic objectives such as standard-plane localization. Our framework integrates real-time ultrasound feedback to enable semantically guided, closed-loop probe manipulation, extending VLA models to imaging-driven medical embodied tasks.

\section{US-VLA}
\subsection{Overview}
As illustrated in Fig.~\ref{fig2}, US-VLA jointly models multi-view RGB observations, clinical instructions, and real-time ultrasound feedback for automated standard-plane acquisition. Wrist- and side-view images with task instructions are encoded by a pre-trained vision–language model, while ultrasound images are processed by USFM \cite{bib33} to capture anatomical and plane-quality information. The ultrasound-aware fusion module integrates both streams through cross-modal attention, and the action expert generates continuous probe commands.

\subsection{Vision–Language Encoding}
US-VLA first performs modality-specific encoding to obtain compact embeddings for downstream alignment and decision making. Specifically, RGB images from the wrist-mounted and side-view cameras are encoded using a SigLIP-based visual encoder \cite{bib34}, producing token sequences $F_W \in \mathbb{R}^{B \times S_{W} \times D_v}$ and $F_S \in \mathbb{R}^{B \times S_{S} \times D_v}$ for the wrist and side views, respectively, where $B$ denotes the batch size, $S_{W}$ and $S_{S}$ are the numbers of visual tokens, and $D_v$ is the visual embedding dimension.

Clinical scanning instructions describing target standard planes are tokenized and embedded by a language encoder, yielding language features $F_L \in \mathbb{R}^{B \times S_{L} \times D_l}$, where $S_{L}$ is the number of language tokens and $D_l$ is the language embedding dimension.

In parallel, we deliberately avoid feeding ultrasound images into the natural-image visual encoder, and instead adopt a universal US foundation model (USFM) \cite{bib33} to mitigate the domain mismatch between natural images and ultrasound imaging. The USFM encoder captures ultrasound-specific anatomical structures and plane quality cues, producing ultrasound features $F_{US} \in \mathbb{R}^{B \times S_{US} \times D_u}$, where $S_{US}$ denotes the number of ultrasound tokens and $D_u$ is the ultrasound feature dimension.

\subsection{Ultrasound-Aware Expert Fusion}
After vision–language encoding, RGB observations and clinical instructions are fused into representations capturing global context and semantic goals. However, they lack fine-grained anatomical and plane-quality cues available only in ultrasound images. We therefore introduce an ultrasound-aware expert fusion module that injects real-time ultrasound feedback by modulating vision–language features through cross-modal attention.


Let $F_{VL} \in \mathbb{R}^{B \times S \times D}$ denote the fused vision--language features produced by the PaliGemma encoder, and let $F_{US} \in \mathbb{R}^{B \times S_{US} \times D}$ denote the ultrasound features encoded by the USFM encoder and projected to the same embedding dimension $D$. Prior to fusion, both feature streams are normalized:
\begin{equation}
\tilde{F}_{VL} = \mathrm{LN}(F_{VL}), \qquad
\tilde{F}_{US} = \mathrm{LN}(F_{US}).
\end{equation}

To selectively inject ultrasound cues into semantic representations, we adopt a cross-modal attention mechanism where vision--language features serve as queries and ultrasound features act as keys and values:
\begin{equation}
Q = \tilde{F}_{VL}W_Q, \quad
K = \tilde{F}_{US}W_K, \quad
V = \tilde{F}_{US}W_V,
\end{equation}
where $W_Q, W_K, W_V \in \mathbb{R}^{D \times D}$ are learnable projection matrices.

Cross-modal attention is computed as:
\begin{equation}
\mathrm{CA}(F_{VL}, F_{US}) =
\mathrm{MHA}(Q, K, V),
\end{equation}
where $\mathrm{MHA}(\cdot)$ denotes multi-head dot-product attention.

The attention output is first combined with the original vision--language features through residual connection:
\begin{equation}
F' = F_{VL} + \mathrm{CA}(F_{VL}, F_{US}),
\end{equation}
and then refined by an expert feed-forward block with another residual connection:
\begin{equation}
F'' = F' + \mathrm{FFN}(\mathrm{LN}(F')),
\end{equation}
where $\mathrm{FFN}(\cdot)$ denotes a two-layer MLP with GELU activation.

Finally, the refined features are projected to the action expert feature width:
\begin{equation}
F_{\mathrm{fused}} = \mathrm{Proj}(\mathrm{LN}(F'')) \in \mathbb{R}^{B \times S \times D_a},
\end{equation}
where $\mathrm{Proj}(\cdot): \mathbb{R}^{D} \rightarrow \mathbb{R}^{D_a}$. The resulting V--L--US fused features are used as the sole input to the subsequent action expert and policy head, enabling closed-loop probe control guided by both semantic goals and ultrasound imaging feedback.

\subsection{Action Expert and Policy Head}
After ultrasound-aware fusion, the resulting multimodal representations encode both task semantics and imaging feedback, which are then used to generate continuous probe control commands. We first extract the last $H_{\mathrm{act}}$ most relevant action-related tokens from the fused sequence,
\begin{equation}
F_{\mathrm{act}} =
F_{\mathrm{fused}}[:, -H_{\mathrm{act}}:]
\in \mathbb{R}^{B \times H_{\mathrm{act}} \times D_a},
\end{equation}
where $H_{\mathrm{act}}=50$ and $D_a=1024$.

These tokens are fed into an action expert network consisting of stacked fully connected layers with LayerNorm and nonlinear activations, yielding expert features $E_{\mathrm{act}} \in \mathbb{R}^{B \times H_{\mathrm{act}} \times D_a}$. In parallel, a lightweight residual branch preserves direct information flow from $F_{\mathrm{act}}$. The two branches are combined via element-wise residual aggregation,
\begin{equation}
F_{\mathrm{out}} = E_{\mathrm{act}} + F_{\mathrm{act}} .
\end{equation}

Finally, a policy head linearly projects the aggregated features to continuous probe actions,
\begin{equation}
A = F_{\mathrm{out}} W_{\mathrm{policy}},
\end{equation}
where $W_{\mathrm{policy}} \in \mathbb{R}^{D_a \times D_{\mathrm{act}}}$. The resulting action sequence is denoted as $A \in \mathbb{R}^{B \times H_{\mathrm{act}} \times D_{\mathrm{act}}}$, representing $H_{\mathrm{act}}$ step-wise continuous probe control commands with $D_{\mathrm{act}}$ action dimensions per step.

\begin{figure}[t]
\centering
\includegraphics[width=0.45\textwidth]{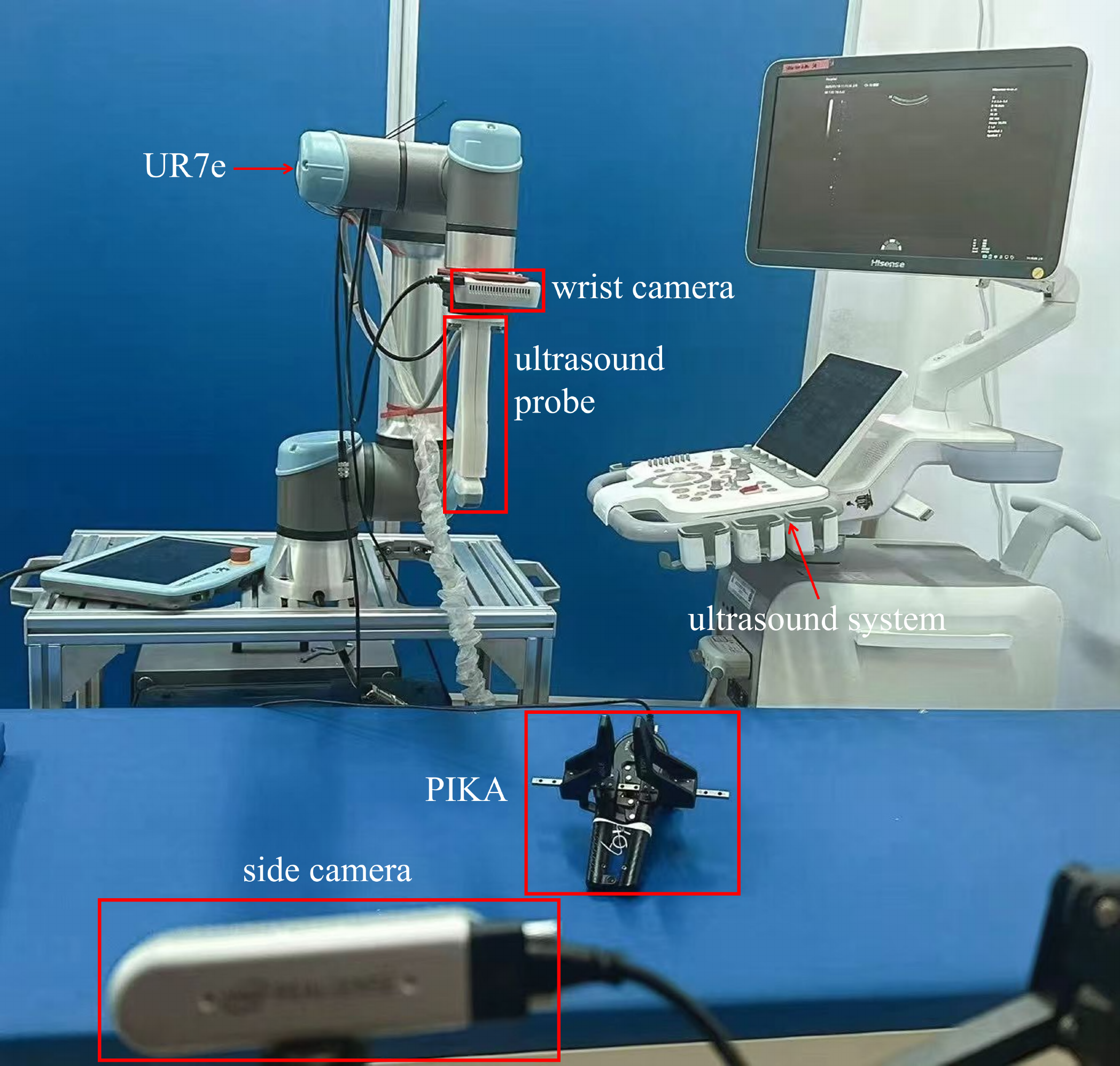}
\caption{Data Collection Setup.}
\label{fig3}
\end{figure}

\begin{figure*}[t]
\centering
\includegraphics[width=0.8\textwidth]{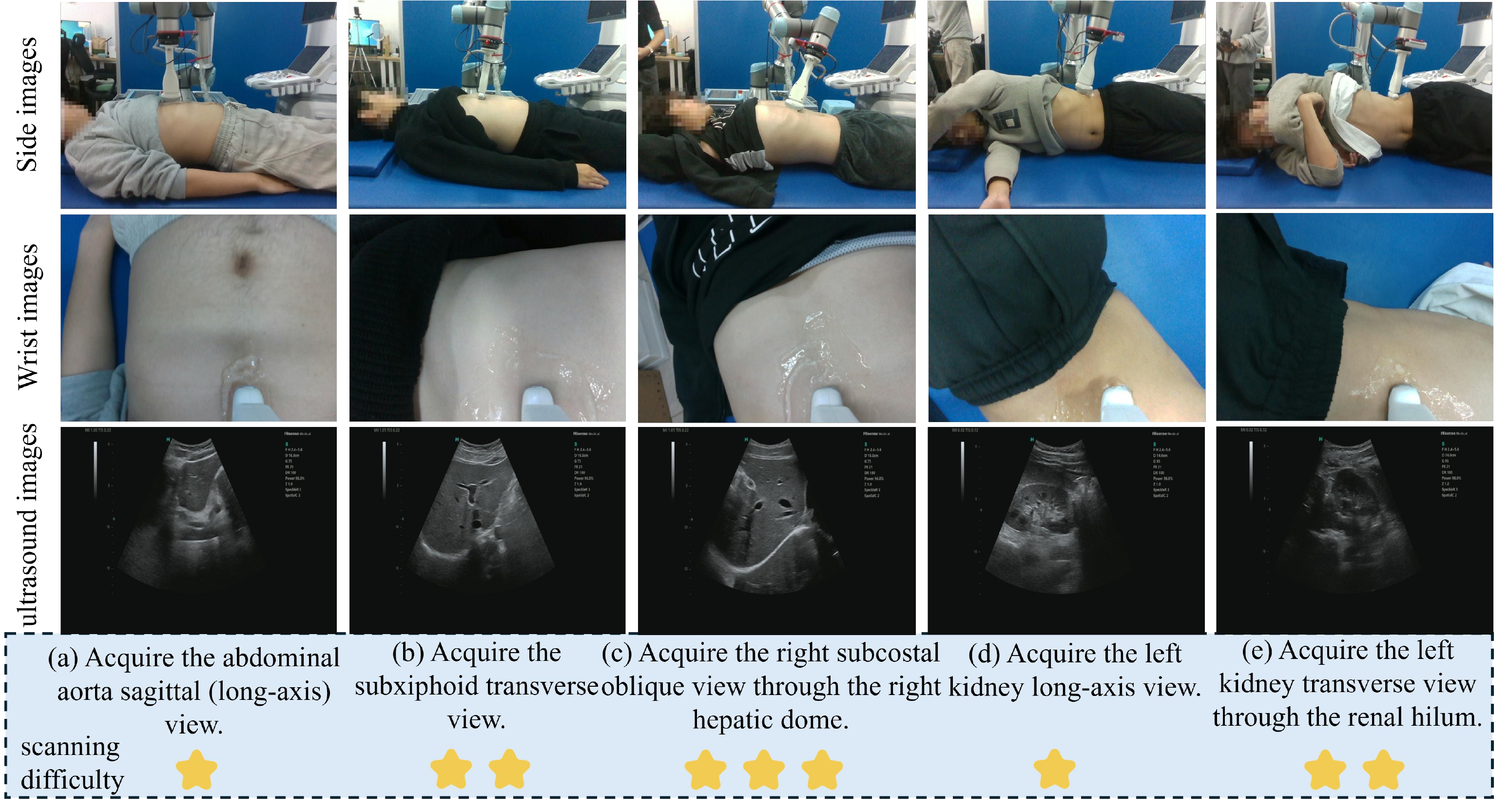}
\caption{Overview of the abdominal ultrasound scanning dataset and task difficulty levels.}
\label{fig4}
\end{figure*}


\section{The Proposed Real-World Dataset}
To train US-VLA, we construct a dedicated dataset for automated ultrasound scanning, in which visual observations, semantic instructions, and ultrasound images are synchronously captured.


As shown in Fig.~\ref{fig3}, data collection is conducted using a UR7e 6-DOF robotic arm. Visual observations are acquired from two RGB-D cameras (Intel RealSense D435, 640×480, 30 FPS) and an ultrasound device, including a side camera and a wrist camera providing egocentric observations. Ultrasound images are captured using a HISENSE Medical HD60 system. All demonstrations are performed by clinically experienced sonographers who remotely control the robotic arm via the PIKA teleoperation system. The complete configuration and setup of the data collection system are provided in the supplementary material.

As shown in Fig.~\ref{fig4}, two expert operators perform five contact-rich ultrasound scanning tasks, including acquiring the abdominal aorta sagittal (long-axis) view, acquiring the subxiphoid transverse view, acquiring the right subcostal oblique view through the right hepatic dome, acquiring the left kidney long-axis view, and acquiring the left kidney transverse view through the renal hilum. During data collection, operators are instructed to complete each target view while exhibiting diverse interaction patterns and strategies. To further increase data diversity, subject postures are adjusted during acquisition to introduce variations in anatomy and conditions.

\textbf{Task 1-Acquire the abdominal aorta sagittal (long-axis) view:} This task mainly evaluates the ability to identify the abdominal aorta. The probe is required to rotate approximately 80$^\circ$--90$^\circ$ to align with the long-axis orientation, resulting in relatively low overall difficulty.

\textbf{Task 2-Acquire the subxiphoid transverse view:} This task requires more precise localization of key imaging regions. The probe must rotate approximately 170$^\circ$--180$^\circ$, imposing higher demands on directional control, and thus exhibits moderate difficulty.

\textbf{Task 3-Acquire the right subcostal oblique view through the right hepatic dome:} This task requires accurate recognition of the liver contour. In addition to a rotation of approximately 80$^\circ$--90$^\circ$, the probe must be further tilted by about 30$^\circ$--40$^\circ$, which demands more complex pose control and spatial understanding, leading to greater overall difficulty.

\textbf{Task 4-Acquire the left kidney long-axis view:} This task focuses on locating the long-axis structure of the kidney. The probe needs to rotate approximately 80$^\circ$--90$^\circ$, while the target anatomy is relatively prominent, resulting in lower difficulty.

\textbf{Task 5-Acquire the left kidney transverse view through the renal hilum:} This task requires identifying the transverse section around the renal hilum. Although the target region is relatively small, no probe rotation is required, and the overall operational difficulty remains low.

The resulting dataset, termed US-VLA-Data, consists of 320 expert demonstration trajectories with approximately 80,000 synchronized timesteps, collected from 8 participants. Each participant performed 5 distinct task types, with each task executed from 8 different positions to ensure diversity and robustness. It is worth noting that the scale of the collected dataset is sufficient to support effective model training. For instance, the ForceVLA-Data dataset used in ForceVLA \cite{bib30}, which involves the same five action categories, comprises a total of 244 trajectories, demonstrating that comparable data scales have been shown to be adequate in related embodied learning tasks.

\begin{figure*}[t]
\centering
\includegraphics[width=0.8\textwidth]{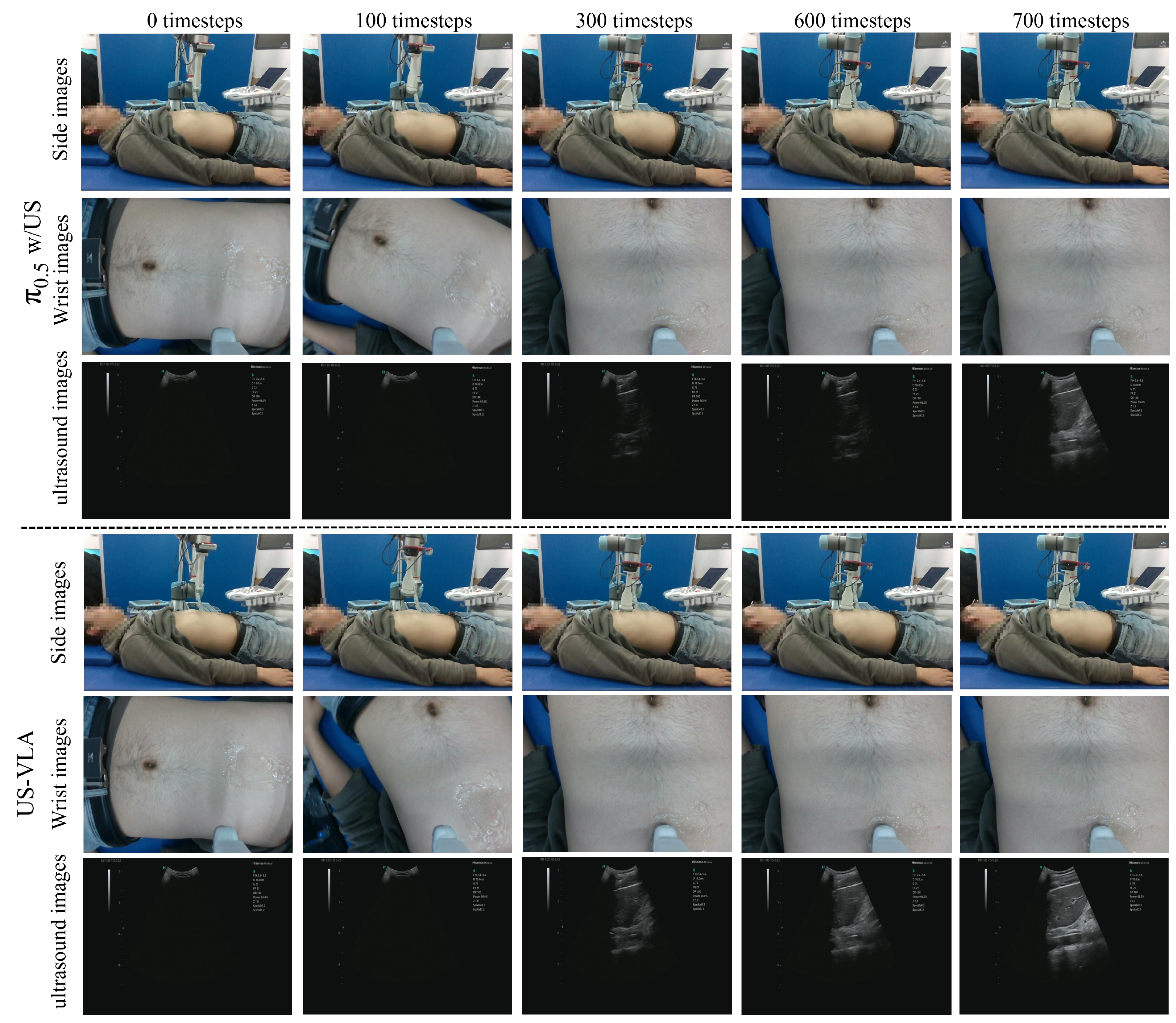}
\caption{Qualitative comparison results of our method and other methods on Task 1.}
\label{fig6}
\end{figure*}

\begin{table*}[t]
\centering
\setlength{\tabcolsep}{5pt}  
\caption{Task performance results on five ultrasound scanning tasks. 
Bold values denote the best performance.}
\label{tab1}
\begin{tabular}{c|cccccc|cccccc}
\hline
\multirow{2}{*}{Model} & \multicolumn{6}{c|}{Success rates $\uparrow$}              & \multicolumn{6}{c}{Scanning timesteps $\downarrow$}          \\ \cline{2-13} 
                       & Task1 & Task2 & Task3 & Task4 & Task5 & Average & Task1 & Task2 & Task3 & Task4 & Task5 & Average \\ \hline
$\pi_0$-base w/o US          &  66.7     &   46.7    &    33.3   &   63.3    &  53.3     &     52.7    &  1329     &   1833    &  1527     &  1498     &    1364   &    1510     \\
$\pi_0$-base w/ US          &  80.0     &   56.7    &   43.3    &   70.0    &   60.0    &   62.0      &   1244    &   1785    &   1439    &   1399    &    1263   &    1426     \\
$\pi_{0.5}$ w/o US            &   73.3    &   53.3    &  40.0    &  66.7     &    56.7   &   58.0      &    1034   &   1706    &   1355    &  1327     &   1115    &    1307     \\
$\pi_{0.5}$ w/ US             &  86.7    &   66.7    &   46.7    &    83.3   &   66.7   &   70.0     &  950     &   1633    &   1300    &   1020    &    972   &    1195     \\

\textbf{US-VLA(Ours)} & \textbf{96.7} & \textbf{76.7} & \textbf{66.7} & \textbf{100} & \textbf{80.0} & \textbf{84.0} &
\textbf{733} & \textbf{1011} & \textbf{882} & \textbf{291} & \textbf{493} & \textbf{682} \\ \hline

\end{tabular}
\end{table*}

\begin{figure*}[!htbp]
\centering
\includegraphics[width=0.79\textwidth]{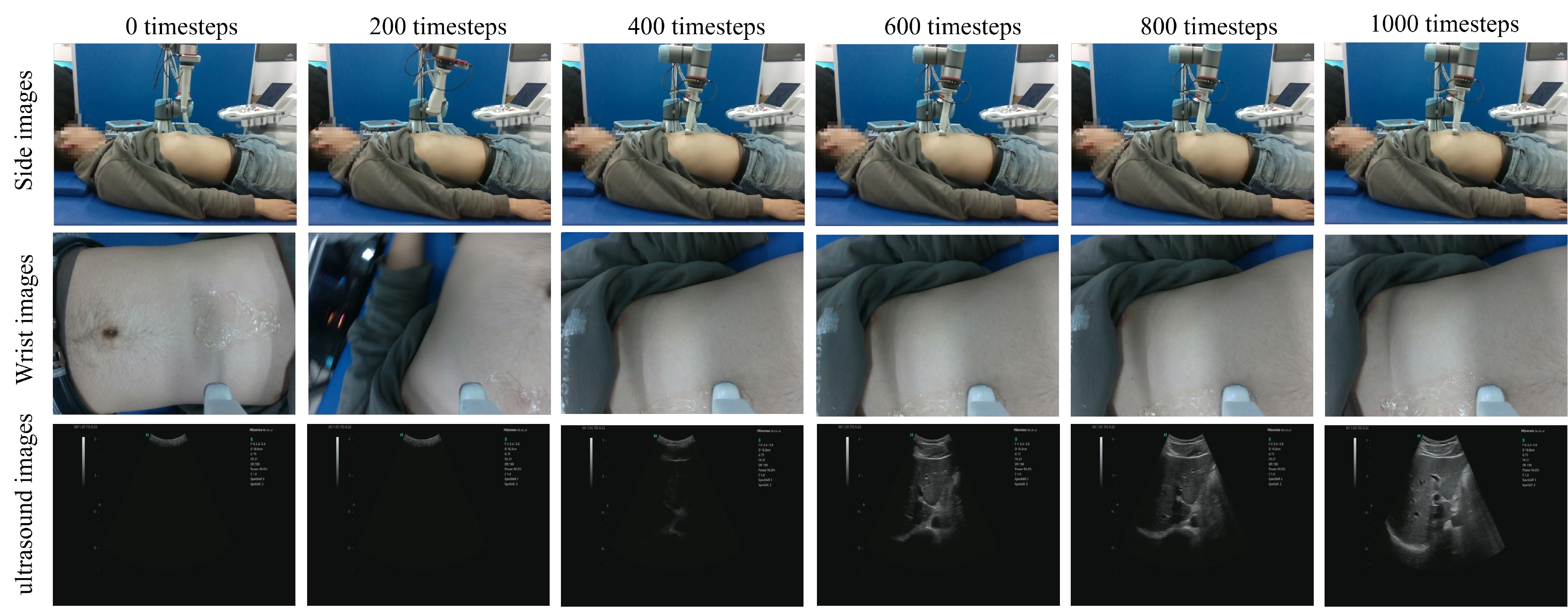}
\caption{Qualitative visualization of our method’s scanning process on Task 2.}
\label{fig10}
\end{figure*}

\begin{figure*}[!htbp]
\centering
\includegraphics[width=0.79\textwidth]{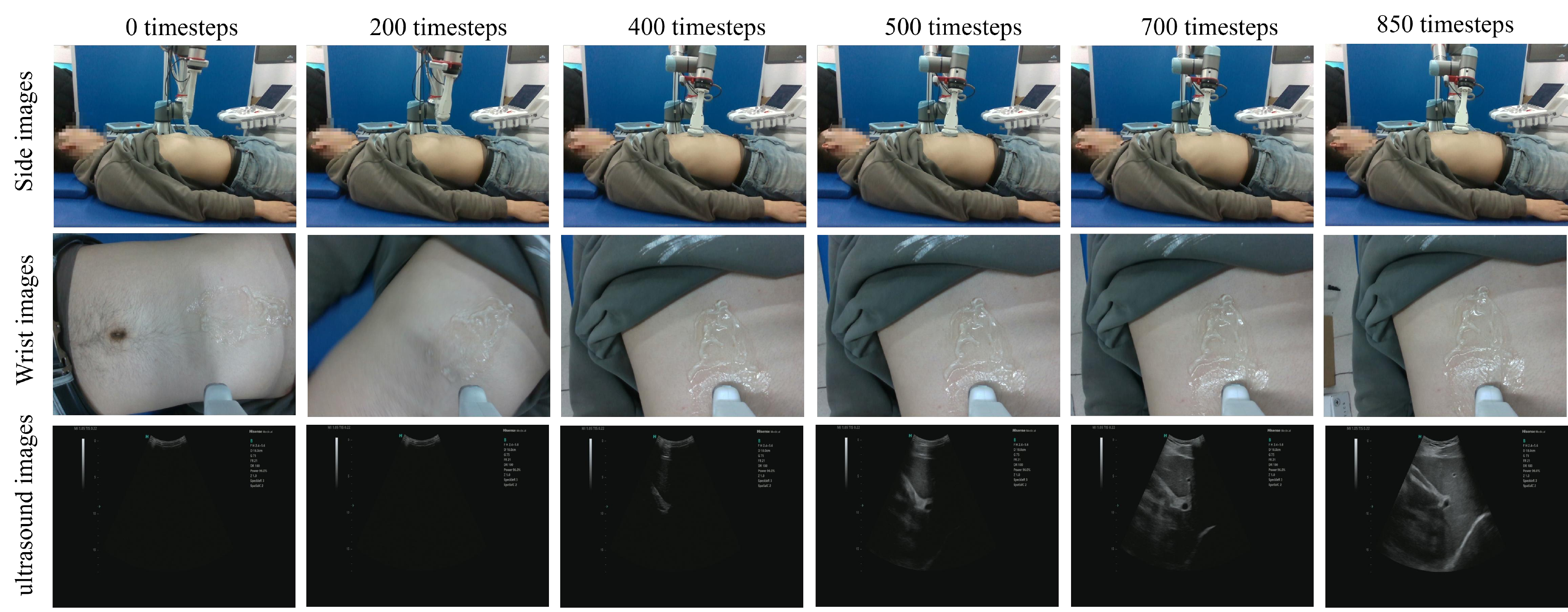}
\caption{Qualitative visualization of our method’s scanning process on Task 3.}
\label{fig11}
\end{figure*}

\begin{figure*}[!htbp]
\centering
\includegraphics[width=0.79\textwidth]{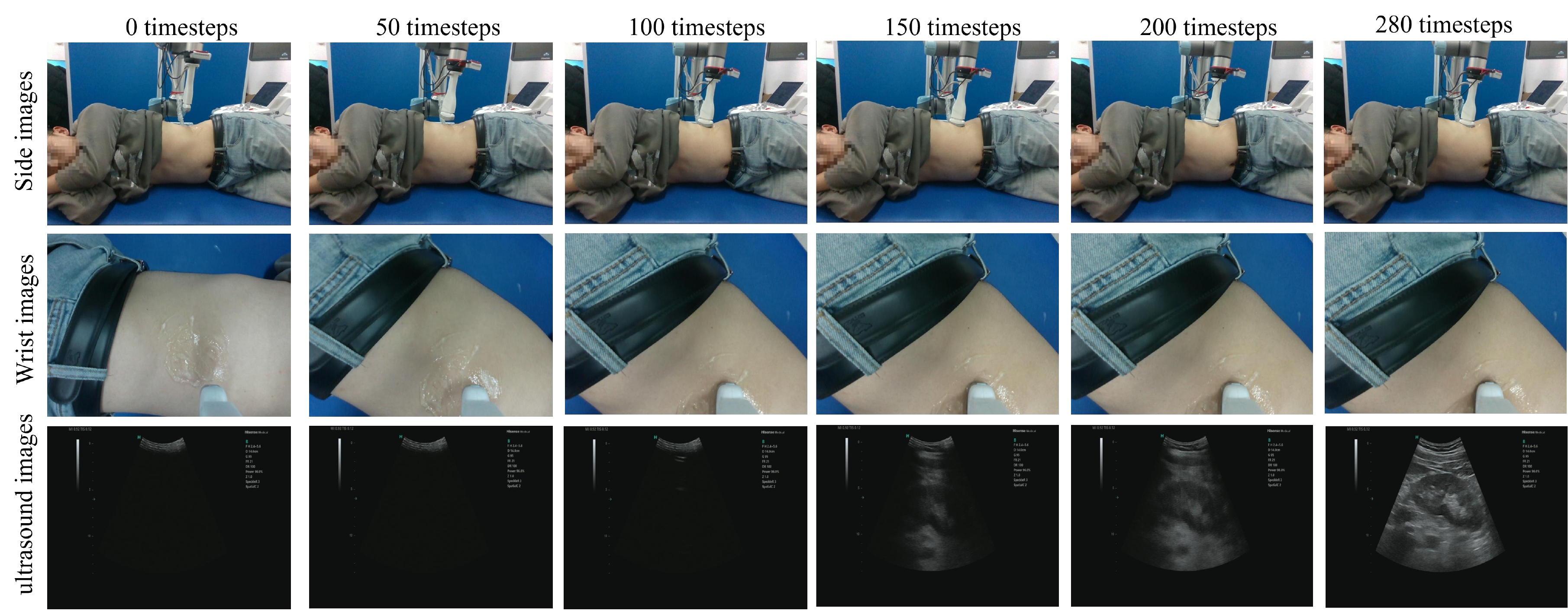}
\caption{Qualitative visualization of our method’s scanning process on Task 4.}
\label{fig12}
\end{figure*}

\begin{figure*}[!htbp]
\centering
\includegraphics[width=0.75\textwidth]{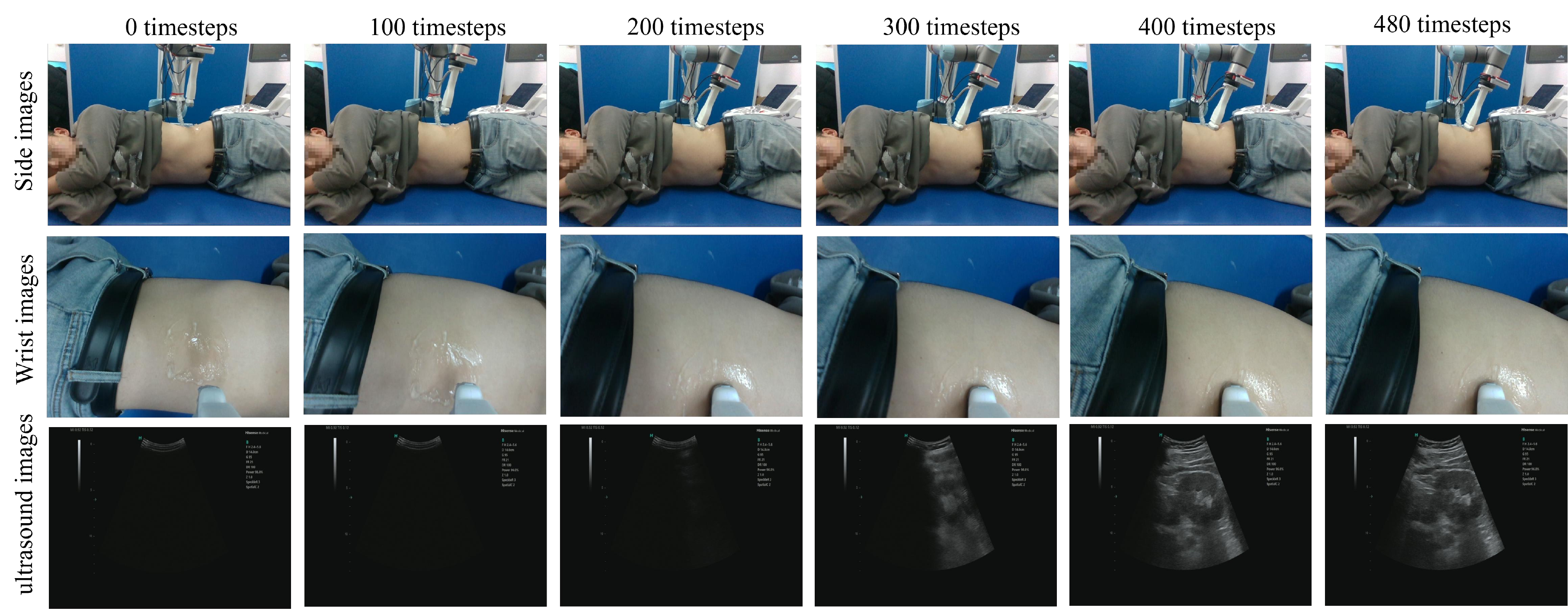}
\caption{Qualitative visualization of our method’s scanning process on Task 5.}
\label{fig13}
\end{figure*}

\section{Experiments}

\begin{table*}[t]
\centering
\setlength{\tabcolsep}{4.5pt}  
\caption{Ablation study of key components in US-VLA.}
\label{tab2}
\begin{tabular}{lll|cccccc|cccccc}
\hline
\multicolumn{1}{c}{\multirow{2}{*}{US}} & \multicolumn{1}{c}{\multirow{2}{*}{USFM}} & \multicolumn{1}{c|}{\multirow{2}{*}{UAEF}} & \multicolumn{6}{c|}{Success rates $\uparrow$}              & \multicolumn{6}{c}{Scanning timesteps $\downarrow$}          \\ \cline{4-15} 
\multicolumn{1}{c}{}                    & \multicolumn{1}{c}{}                      & \multicolumn{1}{c|}{}                      & Task1 & Task2 & Task3 & Task4 & Task5 & Average & Task1 & Task2 & Task3 & Task4 & Task5 & Average \\ \hline
                                        &                                         &                                            &  73.3     &   53.3   &  40.0    &   66.7   &  56.7     &    58.0     &   1034    &   1706    &   1355    &     1327  &   1115    &    1307     \\
$\checkmark$                                       &                                           &                                            &   86.7    &  66.7     &    46.7   &   83.3    &    66.7   &   70.0      &   950    &   1633    &    1300   &   1020    &    972   &    1195     \\
$\checkmark$                                       & $\checkmark$                                         &                                            &   90.0    &   70.0    &   53.5    &   90.0    &   73.3    &     75.3    &    853   &    1549   &   1009    &    708   &   894    &    1062     \\
$\checkmark$                                       & $\checkmark$                                         & $\checkmark$                                          &   96.7    &   76.6    &    66.7   &   100    &  80.0    &    84.0     &   733    &   1011    &    882  &    291   &    493   &    682     \\ \hline
\end{tabular}
\end{table*}

\subsection{Experimental Setups}
\textbf{Evaluation Metrics and Baselines.}
Model performance is primarily evaluated using the task success rate across the five challenging ultrasound scanning tasks, together with timesteps to target. To comprehensively assess the effectiveness of the proposed US-VLA, we compare it with several widely used baselines built upon the state-of-the-art $\pi_{0}$ architecture \cite{bib12}. Specifically, the baselines include $\pi_{0}$-base w/o US (standard $\pi_{0}$ without ultrasound image input), $\pi_{0}$-base w/ US ($\pi_{0}$ with ultrasound signals directly concatenated to the state input), and the corresponding $\pi_{0.5}$variants (with and without ultrasound input) \cite{bib53}, which represent faster execution alternatives under higher control frequencies.

\textbf{Implementation Details.} The model is trained for 30{,}000 steps on 8$\times$ NVIDIA RTX 3090 GPUs using AdamW ($\beta_1=0.9$, $\beta_2=0.95$, $\epsilon=1\times10^{-8}$), with a maximum learning rate of $5\times10^{-5}$ and a batch size of 8. Checkpoints are saved every 5{,}000 steps. US-VLA is trained using 64 expert demonstrations per task and evaluated through 10 independent trials per task on three subjects with different BMI levels to assess stability and generalization under different subject-specific scanning conditions. Both data collection and closed-loop inference operate at 15 Hz.


\begin{figure}[!htbp]
\centering
\includegraphics[width=0.4\textwidth]{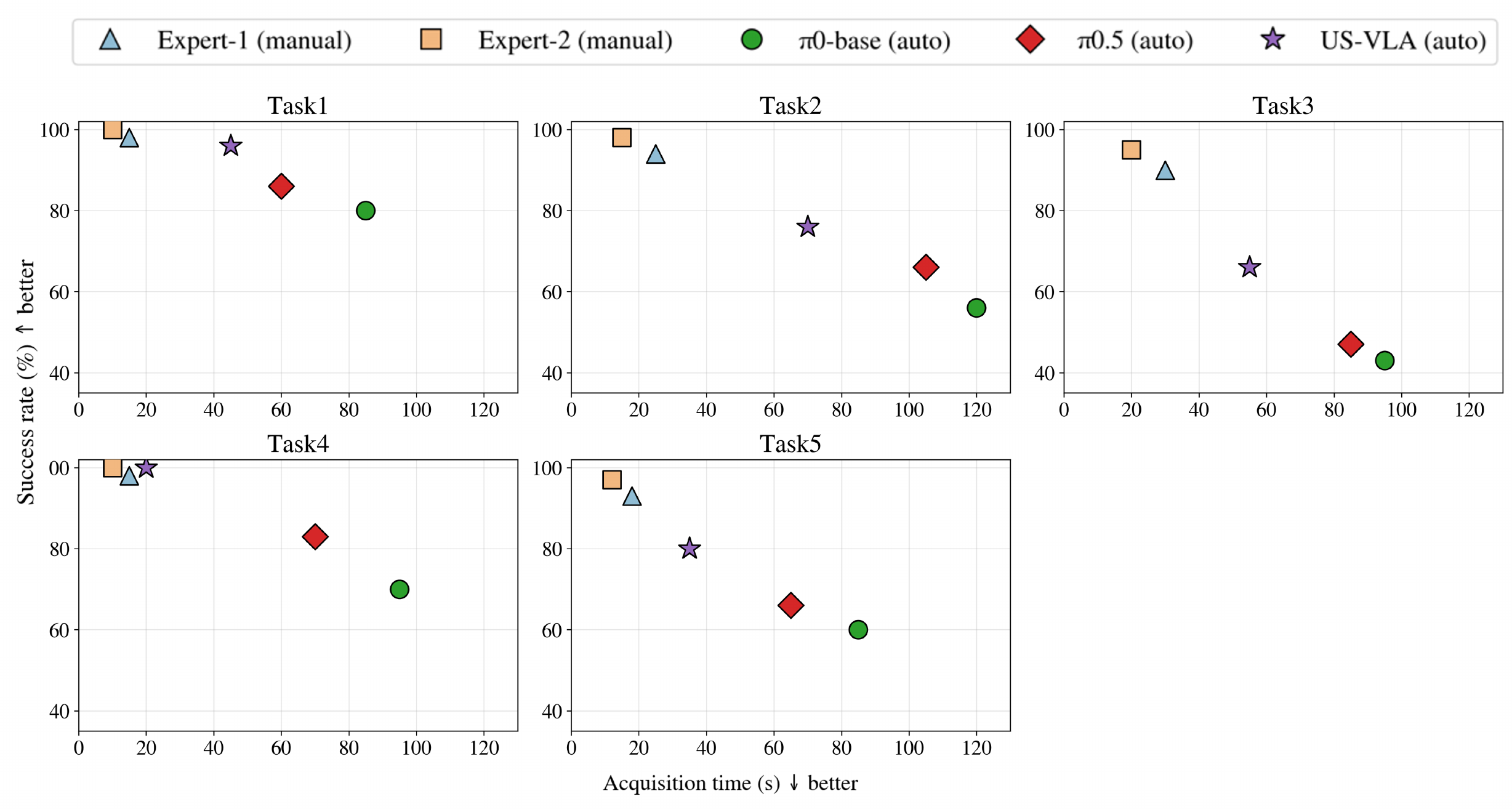}
\caption{Time–success trade-off compared with human experts of different experience levels.}
\label{fig5}
\end{figure}

\subsection{Comparison and Generalization}
\textbf{Quantitative comparison.} As summarized in Table~\ref{tab1}, we compare the success rates and scanning efficiency of different methods across five ultrasound scanning tasks on three held-out subjects with BMI values of 18.1, 22.9, and 26.0, representing diverse body types. US-VLA consistently achieves the best overall performance across all tasks and scanning conditions.

US-VLA attains an average success rate of 84.0\%, outperforming $\pi_{0}$-base w/o US (52.7\%) by 31.3\% and $\pi_{0}$-base w/ US (62.0\%) by 22.0\%. This indicates that simply adding ultrasound input is insufficient without effective ultrasound-aware feature modeling and fusion.

US-VLA also requires only 682 average timesteps to reach the target standard planes, reducing trajectory length by 54.8\% compared with $\pi_{0}$-base w/o US (1510 steps) and by 52.2\% compared with $\pi_{0}$-base w/ US (1426 steps). These results demonstrate that the proposed ultrasound-aware fusion strategy improves target localization and closed-loop probe control, resulting in higher success rates and more efficient scanning trajectories. Participant-level results and scanning videos are provided in the supplementary material for further reference.

\textbf{Qualitative comparison.} As shown in Fig.~\ref{fig6}, we compare the scanning behavior of US-VLA and $\pi_{0.5}$ using synchronized views of probe manipulation, probe–abdomen contact, and ultrasound imaging. US-VLA reaches clinically meaningful standard planes with fewer timesteps, whereas $\pi_{0.5}$ requires longer exploration and exhibits delayed alignment. With ultrasound-aware perception and expert fusion, US-VLA adjusts the probe pose according to real-time ultrasound feedback and follows a coarse-to-fine strategy to refine plane quality. These visualizations further confirm its improved scanning efficiency and stability. Figs.~\ref{fig10}--\ref{fig13} further demonstrate successful liver and kidney scanning performance, indicating strong generalization across diverse tasks.


\subsection{Ablation Studies}
As shown in Table~\ref{tab2}, we perform a systematic ablation study to investigate the individual and combined effects of the three core components in US-VLA on the overall system performance.

\textbf{Effectiveness of Ultrasound Input.}
When only RGB and language inputs are used without ultrasound information, the model achieves an average success rate of 58.0\% across the five tasks, with an average of 1307 scanning timesteps. After introducing ultrasound input, the average success rate increases significantly to 70.0\%, while the average number of timesteps decreases to 1195, indicating that real-time ultrasound feedback indeed plays a crucial and critical role in guiding target-plane search and fine-grained probe pose adjustments.

\textbf{Effectiveness of Ultrasound Feature Modeling (USFM).}
With the inclusion of the USFM for ultrasound-specific representation learning, the performance is further improved to a 75.3\% success rate with only 1062 timesteps, suggesting that ultrasound-aware encoding is crucial for extracting anatomy- and plane-quality–related features beyond generic visual representations.

\textbf{Effectiveness of Ultrasound-Aware Expert Fusion (UAEF).}
With all components enabled, the full US-VLA model achieves the best performance, with the average success rate further improved to 84.0\% and the average scanning timesteps dramatically reduced to 682. This demonstrates that the UAEF module effectively exploits ultrasound features at the decision-making stage and complements visual and language cues, leading to more accurate action generation and substantially shorter search trajectories.

\subsection{Clinical Comparison}

As shown in Fig.~\ref{fig5}, Expert~1 and Expert~2 have one year and three years of clinical ultrasound experience, respectively. Overall, the more experienced expert achieves higher scanning efficiency, while both experts exhibit near-perfect reliability given sufficient adjustment time. In contrast, all learning-based methods perform autonomous closed-loop scanning under limited interaction steps. Among autonomous policies, US-VLA consistently achieves superior time–success trade-off across all tasks, indicating greater acquisition reliability with shorter scanning trajectories compared to baseline methods. These results suggest US-VLA narrows the performance gap between autonomous scanning and manual operation, especially in tasks requiring probe adjustments.

\section{Conclusion}
We introduce US-VLA, a vision–language–action framework for automated ultrasound scanning that explicitly models clinical semantic objectives using real-time ultrasound feedback. An ultrasound-aware expert fusion module fully integrates ultrasound observations with auxiliary contextual information, enabling semantic-driven probe manipulation. We also present US-VLA-Data, a clinically motivated dataset comprising 320 expert scanning trajectories across liver and kidney examinations and five standard planes, representing a range of abdominal ultrasound scanning scenarios. Experimental results show that US-VLA achieves promising and competitive performance in standardized ultrasound scanning tasks. Future work extends the framework to a broader range of organs and clinically defined standard planes, while further incorporating additional sensory feedback, such as force information, to enhance robustness and physical interaction awareness.

\begin{acks}
This work was supported in part by the National Science Foundation of China under Grant 62471448; in part by Shandong Provincial Natural Science Foundation under Grant ZR2024YQ004; in part by TaiShan Scholars Youth Expert Program of Shandong Province under Grant No.tsqn202312109.
\end{acks}

\bibliographystyle{ACM-Reference-Format}
\bibliography{ACM26}

\end{document}